\documentclass{sig-alt-release2}

\usepackage{url}
\usepackage[latin1]{inputenc}
\usepackage{algorithmic}
\usepackage[lined,ruled,boxed]{algorithm2e}
\usepackage{colortbl}
\usepackage{color}
\usepackage[pdftex]{graphicx}

\begin{document}

\title{Multi-Objective Genetic Programming Projection Pursuit \\for Exploratory Data Modeling}

\numberofauthors{2}
\author{
\alignauthor Ilknur Icke{$^1$}\\
        \affaddr{$^1$ The Graduate Center}\\
        \affaddr{City University of New York}\\
        \affaddr{365 Fifth Avenue}\\
        \affaddr{New York, NY 10016}\\
        \email{iicke@gc.cuny.edu}
 \alignauthor Andrew Rosenberg{$^{2,1}$}\\
        \affaddr{$^2$ Queens College}\\
        \affaddr{City University of New York}\\
        \affaddr{65-30 Kissena Blvd.}\\
        \affaddr{Flushing, NY 11367$-$1575}\\
        \email{andrew@cs.qc.cuny.edu}
}

\date{}
\maketitle

\section{Introduction}
For classification problems, feature extraction is a crucial process which aims to find a suitable data representation that increases the performance of the machine learning algorithm. According to the \textit{curse of dimensionality}~\cite{Bellman1961} theorem, the number of samples needed for a classification task increases exponentially as the number of dimensions (variables, features) increases. On the other hand, it is costly to collect, store and process data. Moreover, irrelevant and redundant features might hinder classifier performance. In exploratory analysis settings, high dimensionality prevents the users from exploring the data visually. Feature extraction is a two-step process: feature construction and feature selection. Feature construction creates new features based on the original features and feature selection is the process of selecting the best features as in filter, wrapper and embedded methods~\cite{guyon2006}.

In this work, we focus on feature construction methods that aim to decrease data dimensionality for visualization tasks. Various linear (such as principal components analysis (PCA), multiple discriminants analysis (MDA), exploratory projection pursuit) and non-linear (such as multidimensional scaling (MDS), manifold learning, kernel PCA/LDA, evolutionary constructive induction) techniques have been proposed for dimensionality reduction. Our algorithm is an adaptive feature extraction method which consists of evolutionary constructive induction for feature construction and a hybrid filter/wrapper method for feature selection.

\par\vspace{-10pt}
\section{The Multi-Objective Genetic Programming Projection Pursuit (MOG3P) Algorithm}

We cast the dimensionality reduction task within the genetic programming framework (GP) where the goal is to simultaneously evolve 2 (or 3) data transformation functions that map the input dataset into a lower dimensional representation for visualization. Each function is represented as an expression tree which is made up of a number of base functions over the initial features and represents a 1D projection of the data.

\begin{figure}[h]
\begin{center}
  \par\vspace{-10pt}
   \includegraphics[scale=0.14]{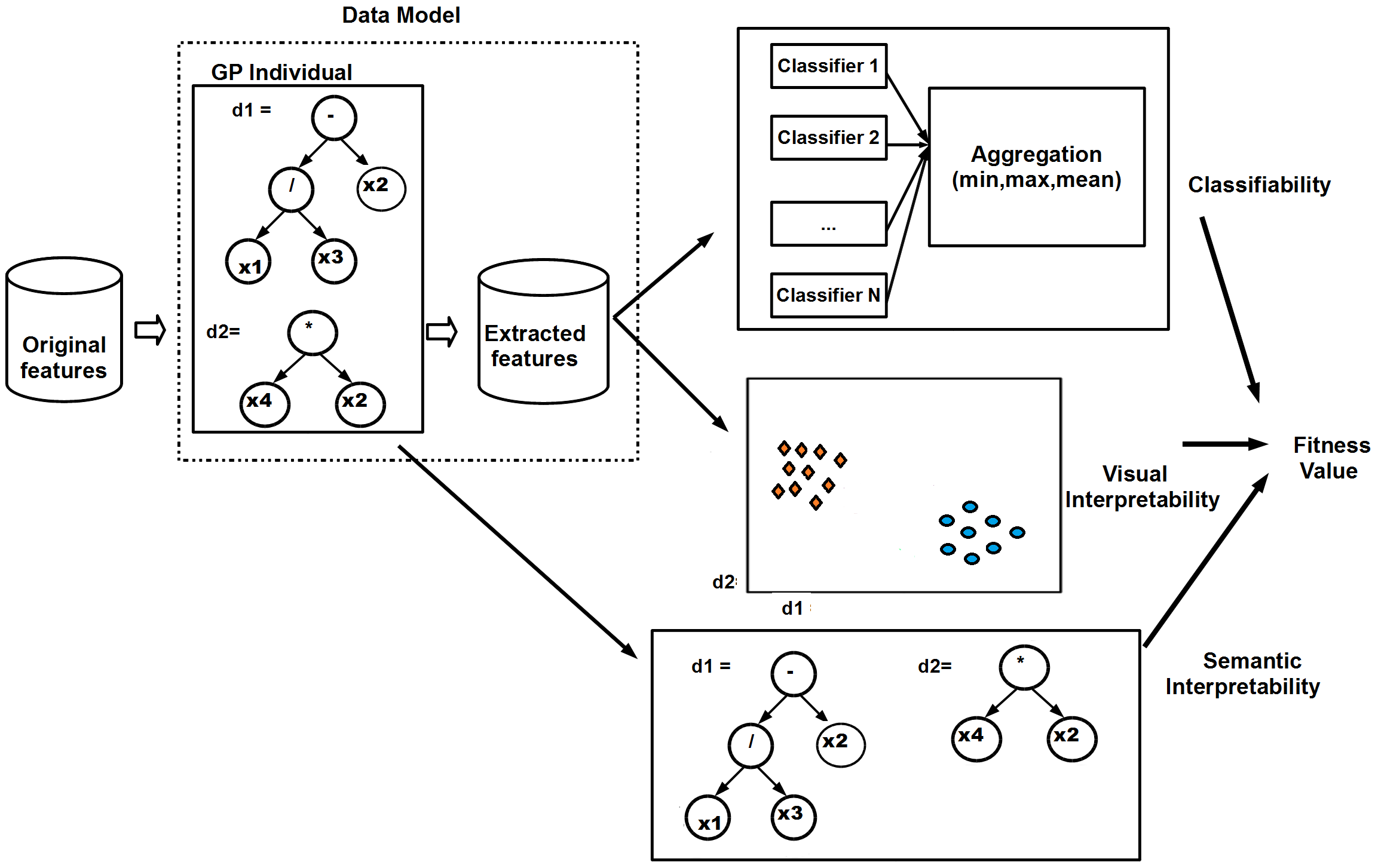}
     \par\vspace{-5pt}
   \caption{\small MOG3P diagram}
    \label{fig:mog3p}
   \end{center}
   \par\vspace{-20pt}
  \end{figure}

  The algorithm is named Multi-Objective Genetic Programming Projection Pursuit (MOG3P) since it searches for \textit{interesting} low dimensional projections of the dataset where the measure of interestingness consists of three equally important objectives (algorithm 1). These objectives are: 1) classifiability: the generated data representation should increase the performance of the learning algorithm(s), 2) visual interpretability: clear class separability when visualized, 3) semantic interpretability: the relationships between the original and evolved features should be easy to comprehend (Figure \ref{fig:mog3p}).

\begin{figure}[h]
\begin{center}
  \par\vspace{-10pt}
   \includegraphics[scale=0.58]{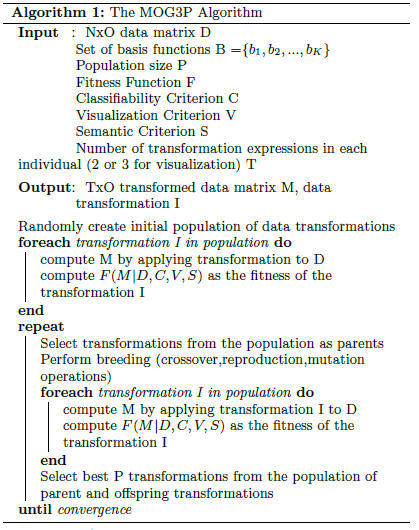}
   \label{algorithm:amog3p}
  \end{center}
   \par\vspace{-20pt}
  \end{figure}

\par\vspace{-15pt}
\section{Experiments}
 In this paper, we report results on two benchmark datasets.
\par\vspace{-5pt}
\begin{table}[h]
\begin{center}
 \scalebox{0.65}{
 \begin{tabular}{|l|c|c|c|c|}
 \hline
  \small Name & $\#$\small features & $\#$\small samples & $\#$\small classes\\
 \hline
 \small Wisconsin Breast Cancer(WBC)~\cite{uci}&\small 9 (ID removed)&\small 683&\small benign:444,malignant:239\\
 \hline
  \small Crabs~\cite{pnnr}&\small 5&\small 200&\small 4 (50 each)\\
 \hline
 \end{tabular}
 }
 \par\vspace{-5pt}
 \caption{{\small Datasets}}
 \label{tbl:datasets}
 \par\vspace{-15pt}
 \end{center}
 \end{table}

We first apply three widely utilized dimensionality reduction techniques on the dataset: principal components analysis (PCA), multidimensional scaling (MDS) and multiple discriminants analysis (MDA). Then we report the 10-fold cross-validation performance of each classifier on these lower dimensional (2D here) representations of the data as well as the original dataset.
PCA and MDA construct new features based on linear transformations of the original features, therefore they can not uncover non-linear relationships. MDS does not construct an explicit mapping between the constructed and original features.

MOG3P is an adaptive algorithm that aims to find the optimal feature representation for the given data. Each candidate data representation is evaluated in a hybrid wrapper/filter manner. Instead of just one classifier we use multiple classifiers. We experiment with WEKA(~\cite{weka}) implementations of the following classifiers: Naive Bayes, Logistic, SMO (support vector machine), RBF Network, IBk (k-Nearest Neighbors), Simple Cart and J48 (decision tree). We examine three ways to compute the classifiability criterion of each individual: 1) maximum, 2) minimum and 3) mean accuracy (10-fold stratified cross-validation accuracy) achieved by any classifier. For visual interpretability, we utilize a measure that is called linear discriminant analysis (LDA) index which is the ratio of the between-group sum of squares to the within-group sum of squares. For semantic interpretability, we consider the total size of the data transformation expressions as the criterion to minimize.

The performance of the MOG3P algorithm is evaluated using a nested 10-fold cross validation scheme in order to assess generalization of the extracted features to unseen data. Table~\ref{tbl:settings} shows the MOG3P settings. Fitness comparisons are made using a pareto dominance based multi-objective optimization method named SPEA2(\cite{spea2}).
\par\vspace{-10pt}
\begin{table}[h]
\begin{center}
 \scalebox{0.6}{
 \begin{tabular}{|l|c|}
 \hline
 Population Size & 400\\ \hline
 Generations & 100 \\ \hline
 Multi objective fitness scheme & SPEA2\\ \hline
 Archive Size & 100 \\ \hline
 Basis functions & $\{+,-,*, protected /,min,max,power,log\}$\\ \hline
 Classifiability Objective (C) & aggregated (min, max, mean) classifier accuracy\\ \hline
 Visualization Objective (V)   & T={2}, LDA index \\ \hline
 Semantic Objective (S)      & Total tree size \\ \hline
 Cross Validation & 10 times 10-fold cross validation (total 100 runs)\\ \hline
 \end{tabular}
 }
 \par\vspace{-5pt}
 \caption{{\small MOG3P Settings}}
 \label{tbl:settings}
 \par\vspace{-20pt}
 \end{center}
 \end{table}

The MOG3P algorithm is a multi-objective machine learning technique which utilizes a population based stochastic optimization approach. Due to this nature, it returns a large number of data models. Each model contains two expressions that construct two new features from the original ones, an LDA index value measuring the visual separation of the class members, as well as training set and test set accuracy values indicating the impact of these new features on each classifier's performance. A good model can be thought of as a model that contains the shortest expressions with smallest LDA index value and highest overall training and test set accuracy on the new feature set. Since there are multiple optimal models, model selection becomes a model mining process. The set of most informative features can be discovered by examining the most frequent features in the optimal models. Moreover, classifier selection can be performed by examining classifier performance across multiple models.

\begin{figure}[h]
      \begin{center}
      \par\vspace{-8pt}
      \includegraphics[scale=0.19]{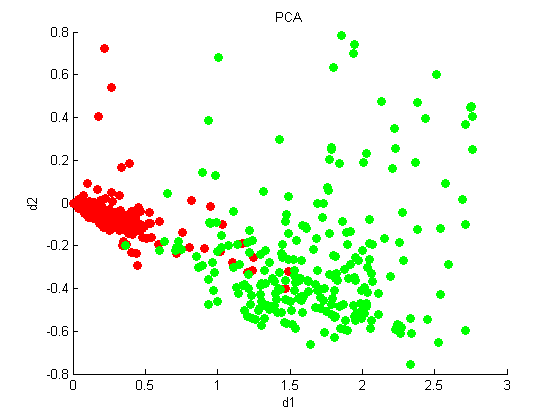}
      \includegraphics[scale=0.19]{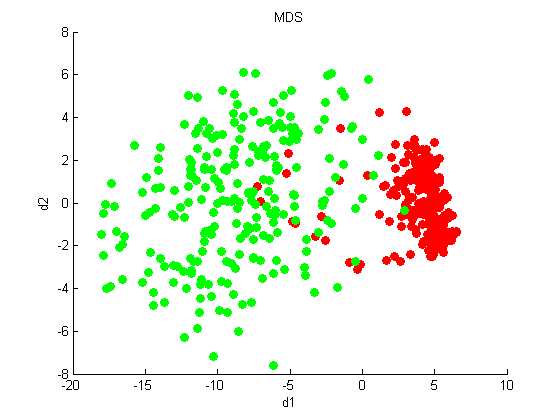}
       \includegraphics[scale=0.19]{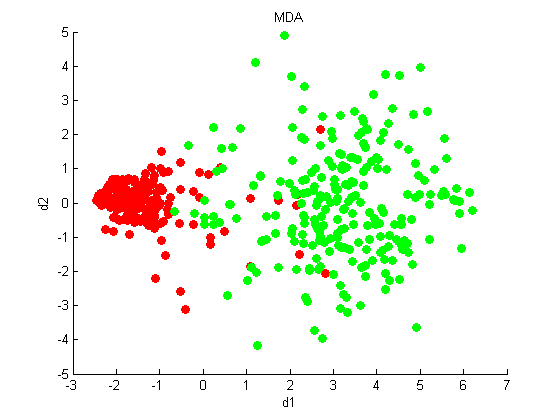}
            \par\vspace{-10pt}
    \caption{\small Visualizations of WBC data(PCA,MDS,MDA)}
      \label{fig:wbcdr}
      \end{center}
      \par\vspace{-20pt}
      \end{figure}

  Figure~\ref{fig:wbcdr} shows visualizations of the WBC data using the standard techniques and table~\ref{tbl:wisconsinresults} shows classifier performances on these lower dimensional representations of the dataset.

      Only MDS and MDA algorithms provide a statistically significant improvement (pairwise t-test) over the original features. 
\par\vspace{-8pt}
\begin{table}[h!tp]
\scalebox{0.48}{
\hspace{0.5cm}
\begin{minipage}[b]{0.5\linewidth}\centering
\begin{tabular}{|l||c|c|c|c|c|c||c|}
\hline
  Classifier & PCA &  MDS &MDA &  All\\
  & (2D) &  (2D) & (2D) &  features  \\
\hline
N. Bayes&96.78 & 97.07&96.78&96.34\\
 \hline
Logistic&96.63&97.07&96.93&96.78\\
 \hline
SMO&96.78&97.07&96.63&97.07\\
 \hline
RBF&96.34&96.63&97.07&95.75\\
 \hline
IBk&95.32&96.49&96.49&95.75\\
 \hline
CART&96.78&97.22&97.07&95.17\\
 \hline
J48&97.22&97.51&96.93&96.05\\
 \hline
\hline
Avg(std)&96.55(0.60)&\textbf{97.01(0.35)}&\textbf{96.84(0.22)}&96.13(0.65) \\
 \hline
 \end{tabular}
\end{minipage}
\hspace{5.2cm}
\begin{minipage}[b]{0.5\linewidth}\centering
\begin{tabular}{|c|c|c|c|c|c||c|}
\hline
   Fitness:minimum & Fitness: maximum & Fitness: mean\\
   MOG3P(2D) & MOG3P(2D) & MOG3P(2D) \\
\hline
   98.21(1.39) & 98.17(1.48) & 98.17(1.48)\\
 \hline
   97.92(1.68) & 97.98(1.73) & 97.94(1.70)\\
 \hline
  97.95(1.66) & 98.04(1.7)  &  97.95(1.61)\\
 \hline
    98.33(1.44)&  98.40(1.43) &  98.38(1.44)\\
 \hline
  98.48(1.44) & 98.58(1.32) &  98.61(1.43)\\
 \hline
  98.30(1.54)  & 98.39(1.52) & 98.26(1.51)\\
 \hline
   98.32(1.50) & 98.29(1.54) & 98.2 (1.54)\\
 \hline
 \hline
 \textbf{98.22(1.53)}&\textbf{98.26(1.54)} &\textbf{98.21(1.54)}\\
\hline
 \end{tabular}
 \end{minipage}
}\\
  \par\vspace{-8pt}
\caption{\small Results on WBC data}
  \par\vspace{-8pt}
\label{tbl:wisconsinresults}
\end{table}

  For all fitness types, the MOG3P algorithm (table~\ref{tbl:wisconsinresults}) finds significantly better data representations (pairwise t-test) that increase accuracy across all classifiers compared to the three standard dimensionality reduction techniques and the original features.

\begin{figure}[h!tp]
  \begin{center}
 \includegraphics[scale=0.35]{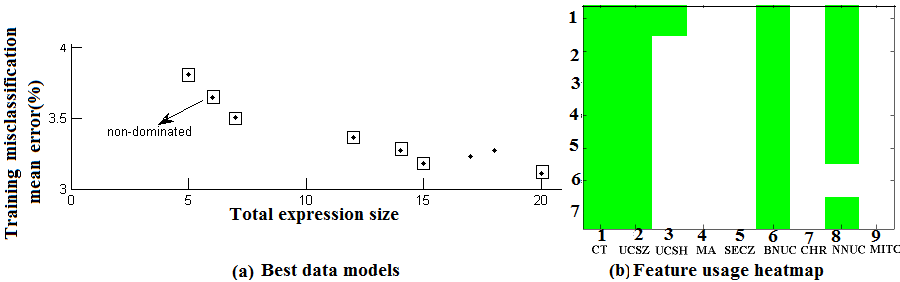}
  \par\vspace{-8pt}
  \caption{\small Results for MOG3P Fitness type: minimum }
  \label{fig:wisconsinresult}
  \end{center}
   \par\vspace{-20pt}
  \end{figure}

Figure~\ref{fig:wisconsinresult}(a) shows the training set error-expression size trade-off for the models with the lowest overall test set error. The best models are marked as non-dominated.
  \par\vspace{-8pt}

 Figure~\ref{fig:wisconsinresult}(b) shows the set of original features that were used by the non-dominated models. These results indicate that only four out of the original nine features were useful for classification.

\begin{figure}[h!tp]
  \begin{center}
  \par\vspace{-8pt}
 \includegraphics[scale=0.19]{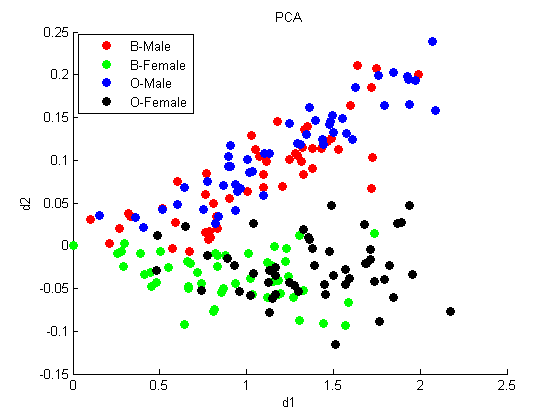}
 \includegraphics[scale=0.19]{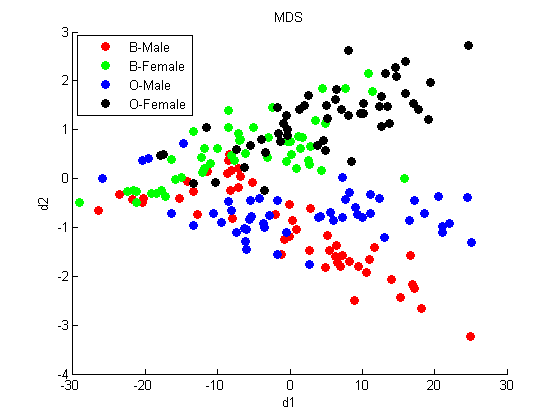}
   \includegraphics[scale=0.19]{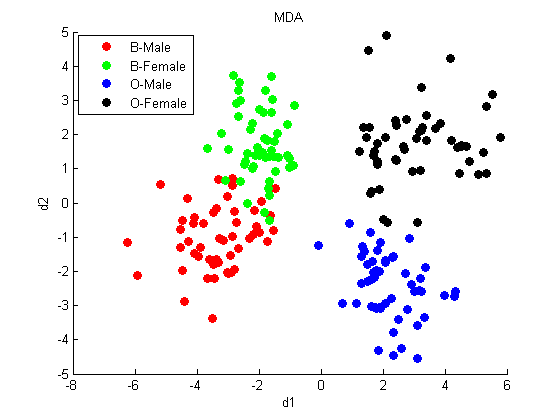}
  \par\vspace{-10pt}
  \caption{\small Visualizations of Crabs data(PCA,MDS,MDA)}
  \label{fig:crabsdr}
  \end{center}
   \par\vspace{-18pt}
  \end{figure}

Figure~\ref{fig:crabsdr} shows visualizations of the Crabs data using the standard techniques and table~\ref{tbl:crabsresults} shows classifier performances on these lower dimensional representations of the dataset.

Only MDA generates a lower dimensional representation that provides a statistically significant improvement (pairwise t-test)  over the original features as well as a visualization that shows clear separation between the classes.

\par\vspace{-5pt}
\begin{table}[h!tp]
\scalebox{0.48}{
\hspace{0.5cm}
\begin{minipage}[b]{0.5\linewidth}\centering
\begin{tabular}{|l||c|c|c|c|c|c||c|}
\hline
  Classifier & PCA &  MDS &MDA &  All\\
  & (2D) &  (2D) & (2D) &  features  \\
\hline

N. Bayes&57.5&67&93.5&38\\
 \hline
Logistic&59.5&63&94.5&96.5\\
 \hline
SMO&54.5&59&94.5&63.5\\
 \hline
RBF&67&69&96&49\\
 \hline
IBk&57&67.5&93&89.5\\
 \hline
CART&57.5&61&94&75.5\\
 \hline
J48&56.5&59&92.5&73.5\\
 \hline
\hline
Avg(std)&58.5(4.03)&63.64(4.19)&\textbf{94(1.16)}&69.36(20.93) \\
 \hline
 \end{tabular}
\end{minipage}
\hspace{4.5cm}
\begin{minipage}[b]{0.5\linewidth}\centering
\begin{tabular}{|c|c|c|c|c||c|}
\hline
  Fitness:minimum & Fitness: maximum & Fitness: mean\\
  MOG3P(2D) & MOG3P(2D) &  MOG3P(2D) \\
\hline
97.8(3.36)  & 97.85(3.12)  & 97.8(2.96) \\
 \hline
 98.1(3) &  98.3(2.77) & 98.1(3.08)\\
 \hline
97.7(3.44) & 97.8(3.43)& 97.8(3.04)\\
 \hline
97.95(3.11) &  97.9(3.35) & 97.95(3.11)\\
 \hline
  97.6(3.3)  &  97.6(3.51) & 97.45(3.52)\\
 \hline
  97.8(3.5) &  97.35(3.99) & 97.8(3.36)\\
 \hline
 97.85(3.5)  & 97.65(3.59)& 97.55(3.59)\\
 \hline
 \hline
 \textbf{97.83(3.31)}  & \textbf{97.78 (3.41)}  & \textbf{97.78(3.24)} \\
\hline
 \end{tabular}
\end{minipage}

}\\
  \par\vspace{-8pt}
\caption{\small Results on Crabs data}
  \par\vspace{-10pt}
\label{tbl:crabsresults}
\end{table}

For all fitness cases, the MOG3P algorithm (table~\ref{tbl:crabsresults}) finds significantly better data representations (pairwise t-test) that increase accuracy across all classifiers compared to the three standard dimensionality reduction techniques and the original features.

\begin{figure}[h!tp]
  \begin{center}
  \par\vspace{-9pt}
 \includegraphics[scale=0.35]{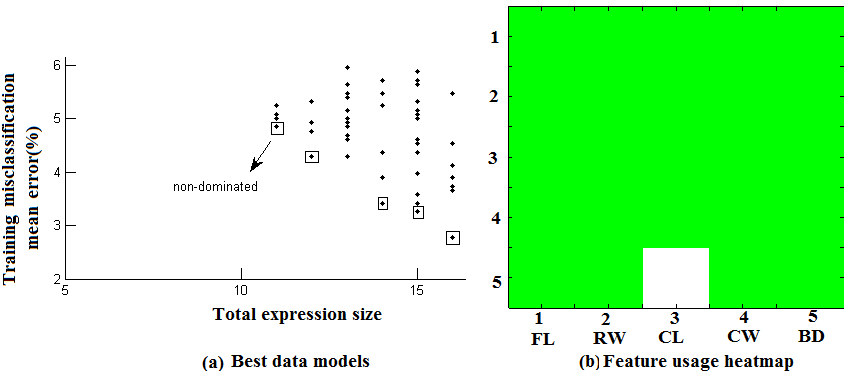}
  \par\vspace{-8pt}
  \caption{\small Results for MOG3P Fitness type: minimum }
  \label{fig:crabsresult}
  \end{center}
   \par\vspace{-20pt}
  \end{figure}

 Figure~\ref{fig:crabsresult}(a) shows the training set error-expression size trade-off for the models with the lowest overall test set error. The best models are marked as non-dominated. The results indicate that all of the original features were necessary for classification.

\par\vspace{-8pt}
\section{Conclusion}
We outline an exploratory approach to data modeling that seeks to simultaneously optimize the human interpretability and the discriminative power. Different measures of interpretability and discriminative power can easily be incorporated into the algorithm in a multi-objective manner without forcing the user to make a-priori decisions on relative importance of these measures. The MOG3P algorithm is a data model mining tool providing the users with multiple optimal models aiming to help them discover the set of most informative features or select a classification algorithm by examining classifier performance across multiple models. Model selection can be performed either by choosing one best model or an ensemble of good models.

\par\vspace{-8pt}
\section{Acknowledgements}
This research was supported in part by a grant of computer time from the City University of New York's High Performance Computing Research Center.
\par\vspace{-9pt}
\bibliographystyle{plain}
\bibliography{literature}

\end{document}